%% file: ms.tex
\documentclass[sigconf]{acmart}
\pdfoutput=1
\usepackage{subcaption} 
\usepackage{listings}   
\usepackage{hyperref}   
\usepackage{graphics}   
\usepackage{balance}    

\definecolor{codegreen}{rgb}{0,0.6,0}
\definecolor{codegray}{rgb}{0.5,0.5,0.5}
\definecolor{codepurple}{rgb}{0.58,0,0.82}
\definecolor{backcolour}{rgb}{0.95,0.95,0.92}

\newtheorem{definition}{Definition}
\newtheorem{proposition}{Proposition}

\AtBeginDocument{%
  \providecommand\BibTeX{{%
    \normalfont B\kern-0.5em{\scshape i\kern-0.25em b}\kern-0.8em\TeX}}}

\copyrightyear{2023}
\acmYear{2023}
\acmConference[AISec '23]{Proceedings of the 16th ACM Workshop on Artificial Intelligence and Security}{November 30, 2023}{Copenhagen, Denmark}
\acmBooktitle{Proceedings of the 16th ACM Workshop on Artificial Intelligence and Security (AISec '23), November 30, 2023, Copenhagen, Denmark}
\acmPrice{15.00}
\acmDOI{10.1145/3605764.3623903}
\acmISBN{979-8-4007-0260-0/23/11}

\begin{document}


\title{When Side-Channel Attacks Break the Black-Box Property of Embedded Artificial Intelligence}

\author{Benoît Coqueret}
\affiliation{%
  \institution{Thales ITSEF}
  \city{Toulouse}
  \country{France}
}
\affiliation{%
  \institution{University of Rennes, INRIA, IRISA}
  \city{Rennes}
  \country{France}
}

\email{benoit.coqueret@thalesgroup.com}

\author{Mathieu Carbone}
\affiliation{%
  \institution{Thales ITSEF}
  \city{Toulouse}
  \country{France}}
\email{mathieu.carbone@thalesgroup.com}

\author{Olivier Sentieys}
\affiliation{%
  \institution{University of Rennes, INRIA, IRISA}
  \city{Rennes}
  \country{France}
}
\email{olivier.sentieys@inria.fr}

\author{Gabriel Zaid}
\affiliation{%
  \institution{Thales ITSEF}
  \city{Toulouse}
  \country{France}}
\email{gabriel.zaid@thalesgroup.com}


\begin{abstract}
  Artificial intelligence, and specifically deep neural networks (DNNs), has rapidly emerged in the past decade as the standard for several tasks from specific advertising to object detection. The performance offered has led DNN algorithms to become a part of critical embedded systems, requiring both efficiency and reliability. In particular, DNNs are subject to malicious examples designed in a way to fool the network while being undetectable to the human observer: the adversarial examples. While previous studies propose frameworks to implement such attacks in black box settings, those often rely on the hypothesis that the attacker has access to the logits of the neural network, breaking the assumption of the traditional black box. In this paper, we investigate a real black box scenario where the attacker has no access to the logits. In particular, we propose an architecture-agnostic attack which solve this constraint by extracting the logits. Our method combines hardware and software attacks, by performing a side-channel attack that exploits electromagnetic leakages to extract the logits for a given input, allowing an attacker to estimate the gradients and produce state-of-the-art adversarial examples to fool the targeted neural network. Through this example of adversarial attack, we demonstrate the effectiveness of logits extraction using side-channel as a first step for more general attack frameworks requiring either the logits or the confidence scores.
\end{abstract}




\maketitle

\input{introduction}
\input{background}
\input{contribution}
\input{experiment}

\input{conclusion}

\begin{acks}
We thank all the anonymous reviewer for their comments, which helped improve the quality of our work. The authors would also like to thank Rémy Baillet and Enzo Rosarini for fruitful discussions about the hardware implementation which improved the quality of our work.
\end{acks}

\bibliographystyle{ACM-Reference-Format}
\balance
\bibliography{bibliographie}

\end{document}

%% file: introduction.tex
\section{Introduction}\label{sec:introduction}
For the past ten years, Deep Neural Networks (DNNs) have improved the automation level in many tasks from image generation \cite{NIPS2014_5ca3e9b1} to object detection for autonomous driving \cite{Redmon2015YouOL}, and Deep Learning (DL) tools are more and more considered by developers. 
With the parallel efforts provided by dedicated hardware designers to make DNNs affordable resource-wise \cite{Shawahna2019FPGABasedAO}, it is not surprising that DL is now thriving in the embedded systems world. 
Such systems are often used in critical environments \cite{9823709, 8625936}, making it vital to be able to assert proof of security and reliability. 
However, DNNs bring both new targets and new threats, thus challenging these requirements. 
For instance, a typical DNN model necessitates a large amount of data and computational time to train and fine tune the model, making it a valuable intellectual property (IP) for the owner and a potential target for an attacker. 
Researches in the security of DNNs have brought to light vulnerabilities both in the physical implementation of these models (hardware level) and in the design of the algorithm (software level). 

Those vulnerabilities can be mainly decomposed into three scenarios.
First, model extraction attacks aim at stealing the trainable parameters (\textit{e.g.}, weights) or the hyperparameters of the model (\textit{e.g.}, neural network architecture) in order to create a clone, compromising the IP. 
Tramèr \textit{et al.} \cite{Tramr2016StealingML} demonstrated the potential of this attack method by taking advantage of prediction APIs in order to copy a small DNN. 
In \cite{Batina2019CSINR}, Batina \textit{et al.} proposed a hardware attack scenario based on side-channel attacks, in order to steal the IP and the inputs of a targeted embedded model.
The practicability of the latter scenario has nevertheless been questioned when an attacker has to deal with real-world and complex DNN implementations \cite{Joud_CARDIS_23}.
A second key consideration for the IP comes from the privacy of the input data during training and inference time. 
Model inversion attacks are designed to predict the inputs of the model, breaking the potential confidentiality of the input data. In their study, Fredriskon \textit{et al.} \cite{Fredrikson2015ModelIA} proved that one could learn sensitive genomic information from the patients, necessitating only the confidence scores of the model. 
Finally, security flaw in DNNs enables an opponent to disrupt the model during inference.
This can be achieved by poisoning the data during training to misclassify or create backdoor during inference, as Yue \textit{et al.} have shown~\cite{Yue2022InvisibleBA}. 
This kind of weakness has also been exploited following a hardware attack strategy, namely, fault injection attack. 
In \cite{8203770}, Liu \textit{et al.} inject faults \textit{via} laser emission to successfully change the output of the embedded network causing a misclassification.
But, this strategy can also be done by perturbing the inference data in an imperceptible way to the human observer but such as it fools the network. 
This scenario is known as evasion or adversarial attacks.
\\

The generation of adversarial examples has been an active research field since its formulation for DL framework in 2013 by Szegedy \textit{et al.} \cite{Szegedy2013IntriguingPO}. 
Since then, several methods, separated by the threat model of the attacker, have emerged. On one end of the spectrum, the attacker is supposed to have full knowledge of the DNN model and its implementation, this is the \textit{white box assumption}. 
In this scenario, several attacks based on optimization problems perform well and allow an attacker to fool the model.
Some of the most impactful methods are FGSM \cite{Goodfellow2014ExplainingAH}, JSMA \cite{Papernot2015TheLO} and C\&W \cite{Carlini2016TowardsET} attack. 
On the other end of the spectrum, the attacker is supposed to have no previous knowledge of the DNN model, this is the \textit{black box assumption}.
In this case, an attacker, referred as Eve in the rest of this work, can only interact with the DNN model in order to capture the predictions it provides.
Therefore, she cannot generate adversarial examples by optimization, as the parameters of the model are unknown. 
A popular solution is to train, using the targeted model as an oracle, a substitute model \cite{Hu2017GeneratingAM} with the goal of copying the original. 
This DNN will allow the attacker to work in a white box setting and use the previously mentioned attack frameworks. 
We argue that this method is far less practical, as it offers lower transfer rate, necessitates higher distortion rate, or both \cite{Chen2017ZOOZO}. A lower transfer rate will defeat the purpose of the substitute network by making it useless, and higher distortion will make the examples more noticeable contradicting with their definition and purpose. The performance might also be impacted by hyperparameters, such as the architecture, of the substitute model, increasing the complexity of the overall attack. Those issues have led researchers to improve the attack by adding assumption to the black box scenario. 
In particular, having access to the logits or the probability vector is one strong hypothesis that is commonly made in several frameworks \cite{Chen2017ZOOZO, Uesato2018AdversarialRA}. 
While generation of more powerful adversarial examples is made possible with this assumption, it is far from a practical one. 
Indeed, one developer can only return the predicted value without providing any access to the logits or the probability distribution. 
If so, the attacker can thus no longer use these methods.

\paragraph{Contributions:}
To solve the black box constraint, we propose a new attack framework, targeting embedded DL systems and combining software and hardware attacks. 
Our method exploits physical side-channel leakages to perform template attacks \cite{ches-2002-635} and deep learning-based side-channel attacks (DLSCA) \cite{Maghrebi_SPACE_2016, Cagli_CHES_2017} to extract the value of the logits vector by targeting the operations during the computation of the softmax layer. 
Since this layer is used to transform, independently from the previous ones, the logits into probabilities, our attack is free from any restricting assumption on the architecture.
Added to the fact that the softmax function is commonly accepted as a standard for the majority of multiclass classifier, it is an ideal target. 
The key idea of our contribution is to add another step to state-of-the-art evasion attacks \cite{Chen2017ZOOZO, Uesato2018AdversarialRA} to produce a generic methodology that allows an attacker to generate adversarial examples in a black box setting, \textit{i.e.}, without prior knowledge on the DNN parameters, including the logits or the probability vector. 
Our main contributions are:
\begin{itemize}
\item  We introduce a new generic attack scenario which extracts the logits from an embedded DL system. Based on this strategy, an attacker takes advantage of side-channel attacks to improve the practicability of state-of-the-art black box adversarial attacks. 
\item We demonstrate the success of our work and its complementarity with state-of-the-art attacks by designing an end-to-end methodology to build powerful adversarial examples by solving the optimization problem proposed by Carlini and Wagner \cite{Carlini2016TowardsET}, via logits extraction and zeroth order optimization.
\item To prove the practicability of our contribution, we implemented a real case scenario on a microcontroller and validated our strategy on a 8-bit quantized DNN using NNOM~\cite{jianjia_ma_2020_4158710}. This experimental setup illustrates the benefits of side-channel attacks to mitigate a black-box scenario.
\end{itemize}

Although this paper focus on the generation of adversarial examples using the ZOO framework, we would like to emphasize that the logit extraction presented here is specific neither to this task (evasion attack), nor to this framework (ZOO). For example, both model extraction methods proposed in \cite{Rolnick2019ReverseengineeringDR} and in \cite{Jagielski2019HighAA} need access to the logit vectors to efficiently steal the model's parameters. With our methodology, an attacker could conduct side-channel attacks to extract the logits in a black-box settings before the application of the methods described in \cite{Rolnick2019ReverseengineeringDR} and \cite{Jagielski2019HighAA}. In other words, the main purpose of this contribution, is to demonstrate the suitability of hardware attacks in making software attacks more practicable under the black-box context by introducing a method to remove a previous assumption, \textit{i.e.}, the access to the logits. \\

This paper is organized as follows. \hyperlink{section.2}{Section 2} presents the notations and the background needed through this study. 
We then describe the principle of our attack (\textit{i.e.}, extraction of the logits) and our end-to-end methodology in \hyperlink{section.3}{Section 3}. 
Finally, we introduce our experimental setup and evaluate our methodology in \hyperlink{section.4}{Section 4} before to conclude in \hyperlink{section.5}{Section 5}.

%% file: background.tex
\section{Background}
\label{Background}
In this section we introduce the necessary background for the comprehension of this paper. We first briefly present deep neural network and quantization. We then describe the two approaches that we will use in our framework: adversarial examples and side-channel attacks.
\subsection{Notation}
Let calligraphic letters $\mathcal{X}$ denote sets, the corresponding capital letters $X$ (resp. bold capital letters) denote random variables (resp. random vectors $\textbf{T}$) and the lowercases $x$ (resp. $\textbf{t}$) denote their realizations.
We will use the following notation $\textbf{T}[i]$ to describe the $i$-th elements of a vector $\textbf{T}$.
Throughout this paper, the function modeled by a DNN is denoted as $f:\mathcal{X} \rightarrow \mathcal{Y}$ and it characterizes its ability to classify a data $X \in \mathcal{X}$ (\textit{e.g.}, an image) over a set of $|\mathcal{Y}|$ classes. The logits, also known as confidence scores, used to perform this classification will be defined by $Z$. To ease the identification of adversarial examples (resp. quantized data), the notation $X^{*}$ (resp. $\tilde{X}$) will be considered.
Finally, the probability of observing an event $X$ is denoted by $\texttt{Pr}[X]$, such that a conditional probability of observing an event $X$ knowing an event $Y$ is denoted $\texttt{Pr}[X|Y]$.

\subsection{Deep Neural Networks}\label{subsec:dnn}
\begin{figure}[ht]
    \centering
    \captionsetup{justification=centering}
    \includegraphics[width=0.47\textwidth]{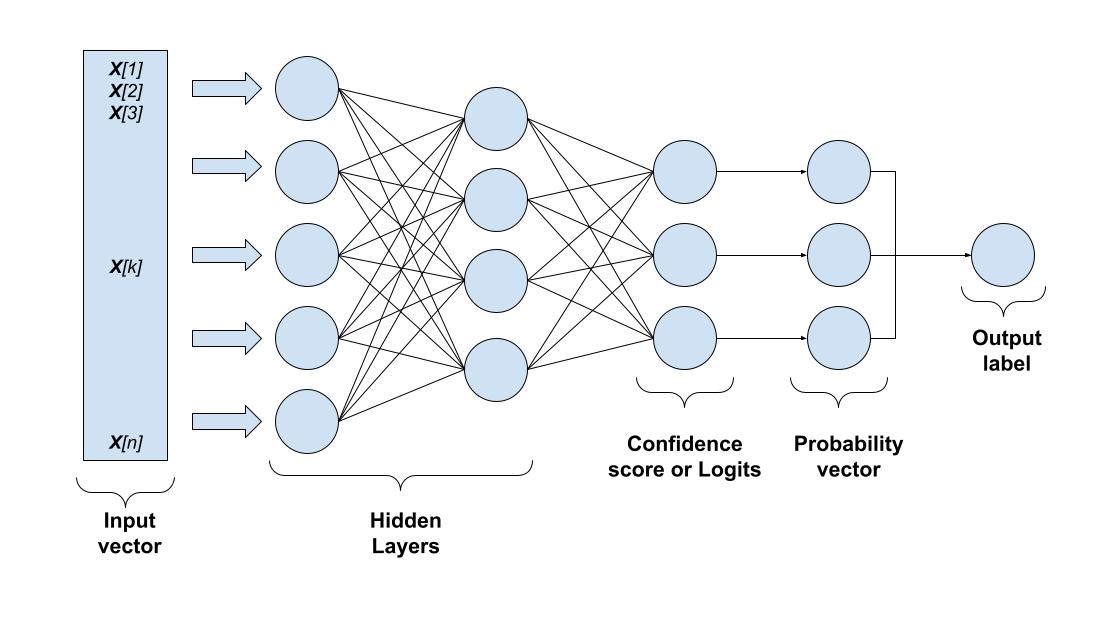}
    \caption{Example of an MLP neural network.} 
    \label{fig:DNN}
\end{figure}

Machine Learning uses statistical tool and optimization algorithms \cite{DL_book} to approximate a function which performs a \textit{classification} or a \textit{regression} task. 
In this paper, a particular focus is brought to the classification task. 
In this setting, a deep neural network is designed to approximate a function $f:\mathcal{X} \rightarrow \mathcal{Y}$ which assigns a unique label $y \in \mathcal{Y}$ to a data $x \in \mathcal{X}$.
To approximate a function $f$ which correctly classifies each input, a set of labeled data $\{(x_{1}, y_{1}), (x_{2}, y_{2}), \ldots, (x_{n},y_{n})\} \in (\mathcal{X} \times \mathcal{Y})^{n}$ is provided in order to monitor its configuration via a loss function and backpropagation. 
This process is known as \textit{supervised learning}, the interested readers may refer to \cite{DL_book} to get further details on the training process of a DNN.
To design the function $f$, a parametric approach which approximates linear and non-linear function is used. 
One solution consists in using a neural network, which is defined as the repetition of a base unit, the \textit{neuron}, stacked by \textit{layers}, forming a combination of simple functions.
An example of the most simple architecture is provided in \autoref{fig:DNN}. It is characterized by a set of fully-connected layers, which corresponds to the multi-layer perceptron (MLP). 
In particular, a neural network is composed by an \textit{input layer} (\textit{i.e.}, the identity over the input data), an \textit{output layer} (\textit{i.e.}, the function which computes the probability vector), and a set of \textit{hidden layers} (\textit{i.e.}, the functions which map the input data to a given class in $\mathcal{Y}$).

Once the function $f$ is correctly designed, the purpose is to select among the possible output classes the most likely class $y$ among the possible outputs included in $\mathcal{Y}$. 
To do so, the last layer is represented by a vector of confidence scores associated with each of the possible class. 
One way to normalize the confidence scores is to perform a softmax function in order to approximate a probability distribution $\texttt{Pr}[Y|X=x]$.
Therefore, the decision-making of the DNN does not only provide the most likely candidate to solve the classification problem, but also the \textit{a posteriori} probability of the remaining $|\mathcal{Y}|-1$ other classes.
This softmax function can be computed such that $s(\textbf{Z}[i]) = \frac{e^{\textbf{Z}[i]}}{\sum_{j=1}^{|\mathcal{Y}|}e^{\textbf{Z}[j]}}$ where $Z[i]$ denotes the score related to the $i$-th class.



\subsection{Quantization}\label{subsec:quantization}

As mentioned in the previous section, DNN models require to keep in memory large amount of data, from weight to intermediate values computed during the inference stage, such as activation value, which can be challenging in resource-constrained environments. One of the most common solution is quantization, which consists in using low precision arithmetic for the model's data. Quantization reduces the number of accesses to the off-chip memory and, as such, the overall power consumption is reduced \cite{Han2015DeepCC}. Two main methodologies for quantization have emerged: Quantization Aware Training (QAT) and Post Training Quantization (PTQ). The first performs the training of the network while taking into account the quantization of the parameters and the latter quantizes the weights once the network is trained.
While numerous complementary adaptation to DL algorithms have been proposed recently \cite{Howard2017MobileNetsEC, Hinton2015DistillingTK, 9043731}, in this paper, we focus on the PTQ solution.

\subsection{Adversarial Examples}
Evasion attack is a type of attack, where a malicious user tries to craft specific inputs such as they are not recognized correctly by the model while appearing normal to a human observer.
These modified inputs are referred to as adversarial examples. Their generation is made by adding imperceptible perturbations (\textit{e.g.}, noise) to the original data in order to force a prediction to a (specific) wrong class. 

\begin{definition}[Adversarial examples \cite{Bernhard2019ImpactOL}]\label{def:ad_ex}
Given a neural network defined by a function $f:\mathcal{X} \rightarrow \mathcal{Y}$, let $X \in \mathcal{X}$ be the original input, $(y, y') \in \{1, \ldots, |\mathcal{Y}|\}^{2}$ the expected label and the targeted adversarial class, then an adversarial example is defined as $\textbf{X}^* = \textbf{X} + boldsymbol{\epsilon}$ with
\begin{equation*}
\begin{split}
    \epsilon &= \underset{\epsilon}{\operatorname{argmin}}||\epsilon||_p,  \\
    s.t f(\text{\textbf{X}}^*) &= y' \text{ (targeted attack)}, \\
    s.t f(\text{\textbf{X}}^*) &\neq y \text{ (untargeted attack)}, \\
\end{split}
\end{equation*}
where $||\epsilon||_p$ is the $l_p$ norm. 
\end{definition}


The generation of adversarial examples is often limited by the level of previous knowledge on the neural network, and in the case where a malicious user has full knowledge over the target, it can be seen as an optimization problem \cite{Szegedy2013IntriguingPO, Carlini2016TowardsET}. 

\paragraph{Carlini and Wagner l2 attack (C\&Wl2).}
In \cite{Carlini2016TowardsET}, the authors tested several objective functions and selected the following formulation for the optimisation problem. 
Given an objective function $g_{obj}$, an input $\textbf{X}$ of dimension $n$ and $\textbf{X}^*$ the targeted adversarial example and $y'$ the adversarial class:  
\begin{equation*}
\begin{split}
\textsf{minimize }& \| \textbf{X}^* - \textbf{X}\|_{2}^{2} + c \times g_{obj}(\textbf{X}^*, y'), \\
&\textsf{s.t  }  \textbf{X}^* \in [0, 1]^{n}, \\
\end{split}
\end{equation*}
Given an hyperparameter $k$ set to 0 in our tests, as suggested by Carlini and Wagner \cite{Chen2017ZOOZO}, the best objective function make use of the logit vector $\textbf{Z} \in \mathbb{R}^{|\mathcal{Y} - 1|}$ and is given by
\begin{equation*}
\begin{split}
g_{obj}(\textbf{X}, y') = \max(\max_{i \neq y'}(\textbf{Z}(\textbf{X})[i]) - (\textbf{Z}(\textbf{X})[y']), - k). 
\end{split}
\end{equation*}

Their framework is still considered to produce state-of-the-art results in a white box setting \cite{Chen2017ZOOZO}, but as such requires full knowledge over the target which is not always a practical solution. On the other hand, in a complete black box setting where the attacker has no knowledge of the DNN parameters, the generated examples are weaker, meaning that they require bigger (and thus more noticeable) transformation of the inputs to produce the same perturbation in the output \cite{Chen2017ZOOZO}. Studies have therefore tried to reduce the gap between the white box generated adversarial examples and the black box ones by using their ability to transfer from one model to another. In this scenario, the malicious user would perform the generation on a substitute model in the white box paradigm, and rely on the possibility of a transfer. Another strategy has been to increase the knowledge of the attacker, by allowing her to have access to the logit or the probability vector of the model, an estimation of the gradient can be made, and it is then possible to generate a white box-like adversarial example.

\paragraph{Zeroth Order Optimization (ZOO)} Chen \textit{et al.} \cite{Chen2017ZOOZO} adapted the loss function from the \textit{C\&Wl2} attack to use the probability vector $\textbf{F}$ instead of the logits. In their method, the objective function to minimize becomes
\begin{equation*}
\begin{split}
g_{obj}(\textbf{X},y') &= \max(\max_{i \neq y'}(\log \textbf{F}(\textbf{X}))[i] - \log (\textbf{F}(\textbf{X}))[y'], - k).
\end{split}
\end{equation*}
They also used finite difference as a proxy for the gradients obtained via backpropagation in a white box settings. Given a scalar $h \in \mathbb{R}$ and a unit vector $\textbf{E} \in \mathbb{R}^n$:
\begin{equation*}
\begin{split}
\frac{\partial f}{\partial \textbf{X}_i} & \approx \frac{f(\textbf{X} + h\textbf{E}[i]) - f(\textbf{X} - h\textbf{E}[i])}{2h}
\end{split}
\end{equation*}
Another similar framework is the SPSA attack \cite{Uesato2018AdversarialRA}, which makes direct use of the logits to approximate the gradient also \textit{via} finite difference. 
Both methods were designed on full-precision network. However, in the context of embedded systems, recent researches have investigated the possibility to use them on quantized neural networks \cite{Bernhard2019ImpactOL}. 
Quantization, especially quantization of the activation values, can deteriorate the generation of adversarial examples by rounding the original and the crafted inputs into the same quantization bucket.
In their study, the authors investigate several frameworks to test their capacity to perform following different metrics~\cite{Bernhard2019ImpactOL}. Their results show that quantization is not by itself a defence against adversarial examples and that all generation frameworks do not react in the same way to quantization.

\subsection{Side-Channel Attacks}\label{subsec:side_channel}
Historically, side-channel analysis (SCA) is a class of cryptographic attack in which an attacker tries to exploit the vulnerabilities of the implementation of a real-word crypto-system for key recovery by analyzing its physical characteristics \textit{via} side-channel traces, like power consumption or electromagnetic emissions. 
During the execution of an algorithm into a crypto-system, side-channel traces record the intermediate variable being processed.
In \cite{Batina2019CSINR}, Batina \textit{et al.} were among the first to transpose this attack to the DNN paradigm by targeting the architecture and the parameters of the model with a Simple Power Attack (SPA) and a Correlation Electromagnetic Attack (CEMA). 
During these, the opponent targets a specific critical variable (\textit{e.g.}, the weights of the DNN) and use a statistical distinguisher to extract it.

\begin{proposition}[Optimal distinguisher \cite{Heuser_CHES_2014}]\label{prop:optim_distinguisher}
The optimal distinguisher $\mathcal{D}_{\text{opt}}$ in SCA context is defined by $\mathcal{D}_{\text{opt}}(\textbf{X}, Y) = \texttt{Pr}[\textbf{X}|Y]$.
\end{proposition}

In other words, the conditional probability mass function (\textit{pmf}) is the best model we can build in the SCA context.
However, such probability remains unknown for the attacker.
Therefore, an approximation of such unknown conditional \textit{pmf} is required.
To conduct such approximation, the best solution is to perform a \textit{profiled attack}.
In this scenario, it is assumed that the attacker has an access to a copy of the victim's device with a full knowledge to the targeted component. 

\begin{definition}[Targeted Device]\label{def:targeted_device}
In the context of profiled attack, the targeted device $\mathcal{D^*}$ corresponds to a system implementing an insecure function on a specific hardware for which the attacker has a model allowing her to infer information on sensitive variables.
\end{definition}
\begin{definition}[Open Device]\label{def:open_device}
Given a target device $\mathcal{D^*}$, an open device $\mathcal{D}$ refers to a copy of the targeted device where the attacker has access to both the physical traces $\mathcal{X}_{p} = \{\textbf{x}_{1}, \ldots, \textbf{x}_{N_{p}}\}$ and the corresponding labels $\mathcal{Y}_{p} = \{\textbf{y}_{1}, \ldots, \textbf{y}_{N_{p}}\}$ to build a sub-optimal solution to \hyperlink{proposition.1}{Proposition 1}. 
\end{definition}

This kind of attacks is performed in two phases, \textit{i.e.}, profiling and exploitation phases:
\begin{enumerate}
    \item In the profiling phase, the attacker gathers leakage information from an open copy $\mathcal{D}$ of the target device $\mathcal{D^*}$, embedding the AI, to extract a sensitive intermediate variable $y$. For this purpose, a set of $N_{p}$ profiling physical traces $\mathcal{X}_{p} = \{\textbf{x}_{1}, \ldots, \textbf{x}_{N_{p}}\}$ are collected encompassing time samples related to the manipulation of $Y$. Then, an estimation of multivariate class-conditional probability distributions of $\mathbb{X}$ is computed ($\texttt{Pr}[\mathbb{X}|Y = y])$) over all classes $y \in \mathcal{Y}$.
    \item In the exploitation phase, the attacker gathers leakage information from the target device $\mathcal{D^*}$, embedding the AI, focusing on the identical, but now unknown, class $y$. Considering the adversary has the set of probability distribution estimations at disposal (precomputed during profiling phase), the attack is mounted against a set of $N_{a}$ exploitation traces $\mathcal{X}_{a} = \{\textbf{x}_{1}, \ldots, \textbf{x}_{N_{a}}\}$. This attack is based on \textit{maximum a posteriori} (MAP) rule by estimating likelihood over all classes in $\mathcal{Y}$. Then, by looking at the most likely candidate via maximization, \textit{i.e.}, $\underset{y \in \mathcal{Y}}{\texttt{argmax}} \sum_{i=1}^{N_{a}} \log (\texttt{Pr}[\textbf{X} = \textbf{x}_{i} | Y = y])$, an estimation is made on the correct class $y$. For computation stability, one usually computes \textit{log-likehood}.
\end{enumerate}

To find a sub-optimal solution to \hyperlink{proposition.1}{Proposition 1}, two approaches can be considered namely, template attacks \cite{ches-2002-635} or DLSCA \cite{Maghrebi_SPACE_2016, Cagli_CHES_2017}.
Whereas the template attacks approximate the unknown conditional \textit{pmf} $\texttt{Pr}[\textbf{X}|Y]$, the DLSCA uses supervised learning to estimate the unknown conditional \textit{pmf} $\texttt{Pr}[Y|\textbf{X}]$. 
Even if both solutions are equivalent up to a constant in SCA\footnote{In the SCA context, if $Y$ is uniformly distributed over $\mathcal{Y}$, $\texttt{Pr}[\textbf{X}|Y]=\frac{\texttt{Pr}[Y|\textbf{X}]\cdot\texttt{Pr}[\textbf{X}]}{\texttt{Pr}[Y]} = \epsilon \cdot \texttt{Pr}[Y|\textbf{X}]$ where $\epsilon$ does not depend on $Y$.}, some benefits and limitations can be expressed for each method. 

\paragraph{Template attacks.}
Therefore, to approximate the unknown conditional \textit{pmf} $\texttt{Pr}[\textbf{X}|Y]$, the profiling phase consists in estimating the mean vector $\boldsymbol{\mu}_{y}$ and the covariance matrix $\Sigma_{y}$ for each class $y \in \mathcal{Y}$ from a set $\mathcal{X}_{p}$ of profiling physical traces.
Then, the sub-optimal solution of \hyperlink{proposition.1}{Proposition 1} is defined by a Gaussian probability density function (\textit{pdf}) with parameters $\boldsymbol{\mu}_{y}$ and $\Sigma_{y}$.
However, the huge dimensionality\footnote{In real-world use-case, the dimensionality of physical traces can easily exceed $10^{7}$.} of $\textbf{X}$ can sometimes make the estimation of \hyperlink{proposition.1}{Proposition 1} a complex task.
Therefore, a popular solution consists in selecting a small portion of the physical traces based on some statistical tests \cite{Choudary_CARDIS_2014} (\textit{e.g.} SNR \cite{DPA_book}, difference of means \cite{ches-2002-635}, T-Test \cite{Becker2013TestVL}) or dimensionality reduction techniques (\textit{e.g.} PCA \cite{Archambeau_CHES_2006}, LDA \cite{Choudary_CARDIS_2015}, KDA \cite{Cagli_CARDIS_2017}).
In this paper, only the SNR metric will be considered for selecting a small portion of the physical traces.
Another limitation of the template attacks relies on the Gaussian assumption.
Indeed, when the underlying Gaussian assumption does not hold, estimating $\texttt{Pr}[\textbf{X}|Y]$ is known
to be hard in practice \cite{Bruneau2017OptimalSA}.
This has led the side-channel community to find alternatives based on Machine Learning techniques.

\paragraph{Deep learning-based side-channel attacks (DLSCA)}
Over the past few years, deep learning approaches have been investigated in the SCA context to mitigate the impact of some countermeasures, namely desychronization \cite{Cagli_CHES_2017, Masure_ESORICS_2020} and masking \cite{Maghrebi_SPACE_2016, ASCAD}.
Similarly to the template attacks, two phases are required. 
First, the profiling phase is applied to approximate the true unknown conditional probability $\texttt{Pr}[Y|\textbf{X}]$ by following the strategy detailed in \hyperlink{subsection.2.2}{Section 2.2}.
Indeed, a DNN is designed by the attacker to select a function $f$ which maps a physical trace to a sensitive variable $y \in \mathcal{Y}$. 
The structure-\textit{agnostic} property of the MLP is one of its main advantages.
Indeed, to consider it, no particular assumption has to be made on the data structure. 
In a side-channel context, this is beneficial to automatically find a function which is not limited to the Gaussian assumption.


%% file: contribution.tex
\section{Side-Channel Attacks Make Adversarial Examples More Practicable}\label{subsec:contribution}
This section introduces a new black-box attack scenario against embedded neural networks.
It takes advantage of hardware side-channel attacks to demonstrate the ability of an attacker to extract sensitive information (\textit{i.e.}, logits) which are usually considered as known.
As this assumption is usually not practicable, a new contribution must be introduced to extract those logits.
This section demonstrates the synergy between hardware and software attacks to defeat the security of an embedded neural network.
\hyperlink{subsection.3.1}{Section \ref{subsec:attack_principle}} describes the attack strategy by introducing our contribution, as well as which threat models can be considered in such scenario.
Finally, \hyperlink{subsection.3.2}{Section 3.2} details our contribution through the identification of the attack path, and the practical recommendations that are needed to perform it. 

\subsection{Methodology}\label{subsec:attack_principle}
Our contribution provides an independent method to break free from assumption traditionally restricting the state-of-the-art frameworks for black box generation of adversarial examples. 
For example, two widely adopted solutions, namely ZOO \cite{Chen2017ZOOZO} and SPSA \cite{Uesato2018AdversarialRA} attacks, require access to the logits or the probability vector, while being under the assumption of the black box.
Following the black box definition of this paper, such strategy could not be conducted by any attacker as she would only have access to the predicted label included in $\mathcal{Y}$.
Therefore, the requirements of ZOO \cite{Chen2017ZOOZO} and SPSA \cite{Uesato2018AdversarialRA} are not fulfilled.
In this context, our contribution takes advantage of hardware attacks to complete state-of-the-art adversarial example generation, and provides an access to the logits and the probability vector in a black box context.
This forms an overall assumption-free attack (\textit{e.g.}, no previous knowledge on the DNN's hyparemters or parameters is required, nor access to the logits, confidence scores or even output label).

The first step of our attack is to design a sub-optimal solution of \hyperlink{proposition.1}{Proposition 1}; this is a crucial step that needs an access to the copy of the targeted device. As explained, in Definition \ref{def:open_device}, $\mathcal{D}$ enables the malicious user to create a statistical distinguisher, which will be used to attack the softmax function of the targeted device $\mathcal{D}^*$ running the neural network. This corresponds to the profiling phase.

\begin{figure}[t!] 
\begin{subfigure}{0.23\textwidth}
\includegraphics[width=\linewidth]{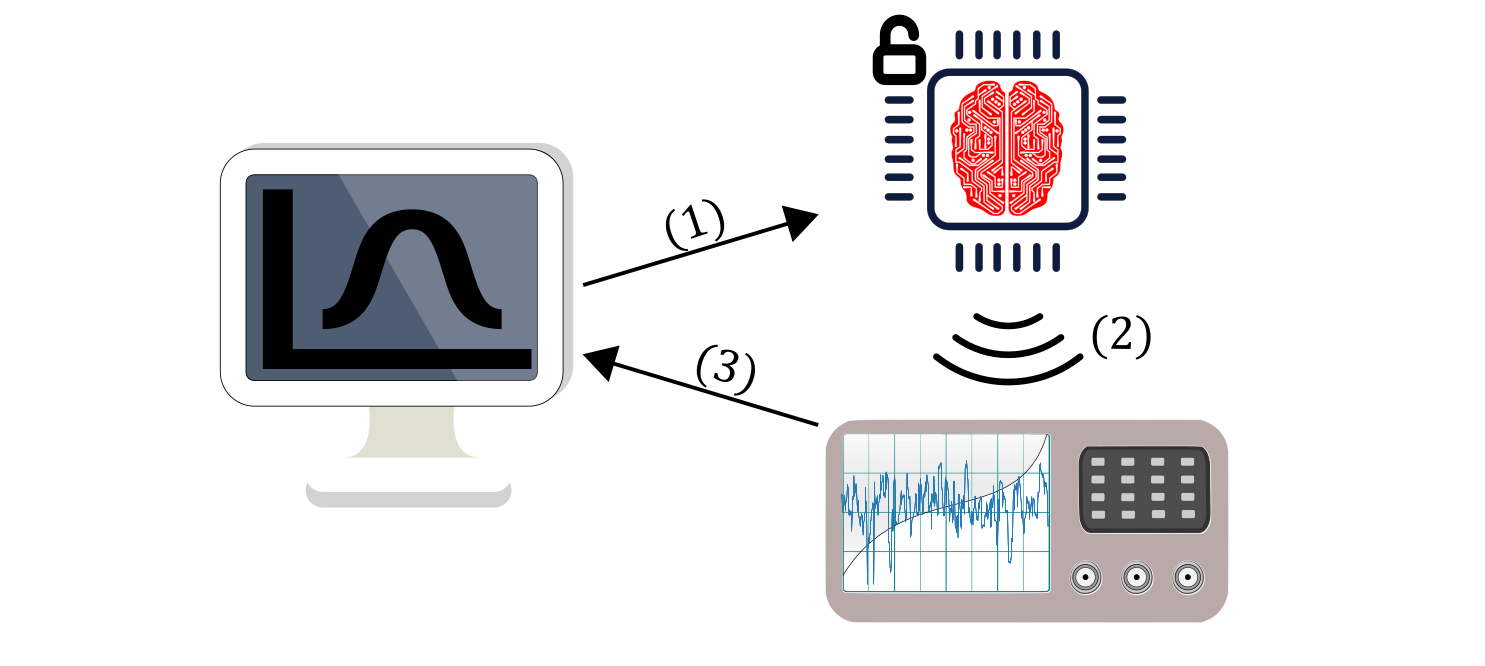}
\caption{Profiling phase} \label{fig:a}
\label{fig:profiling_phase}
\end{subfigure}\hspace*{\fill}
\begin{subfigure}{0.23\textwidth}
\includegraphics[width=\linewidth]{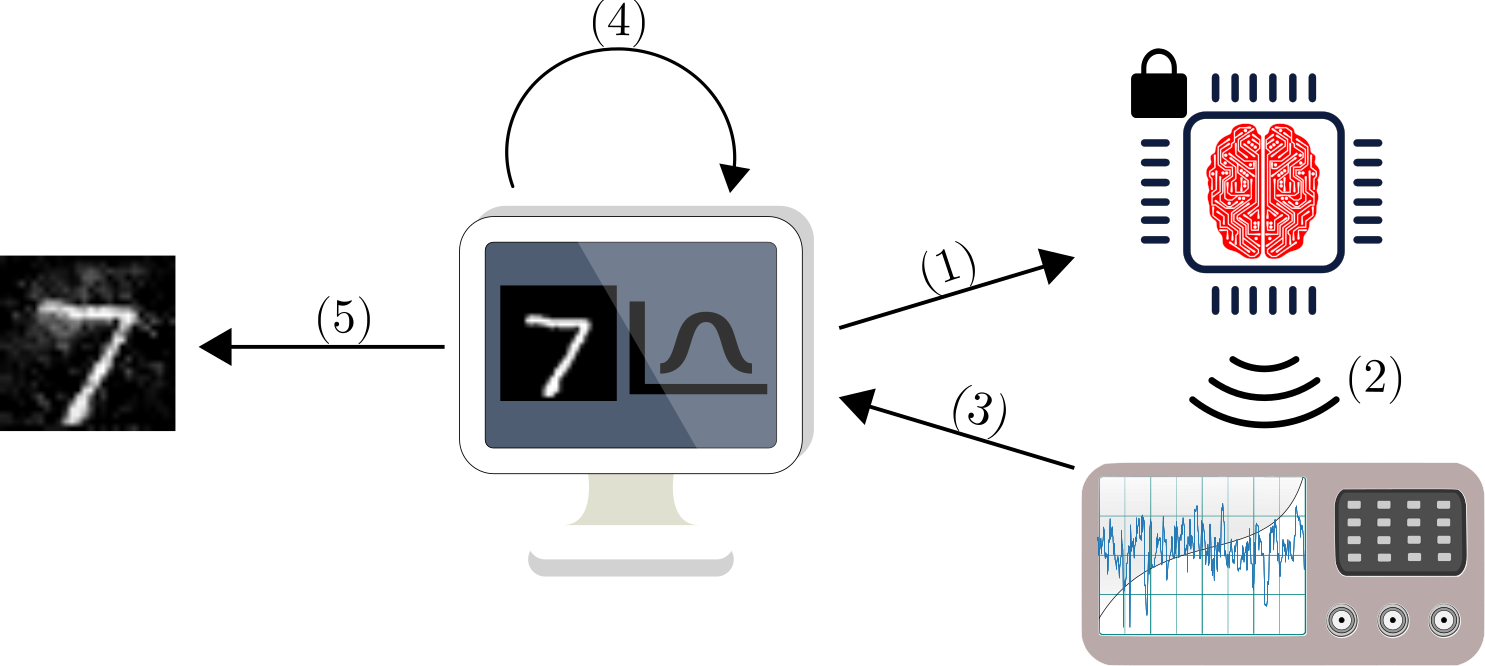}
\caption{Attack phase} \label{fig:b}
\end{subfigure}
\caption{Schematic description of the profiling and attack phases of our attack.}
\label{fig:attack_phase}
\end{figure}

The process, illustrated in \autoref{fig:profiling_phase}, can be detailed as follows:
\begin{itemize}
\item (1) A set of random input data, for which the logits are known, are sent to the embedded neural network for prediction;
\item (2) During prediction, the attacker captures the related side-channel traces which contain information on the manipulated logits;
\item (3) The attacker creates a sub-optimal solution of \hyperlink{proposition.1}{Proposition 1} (\textit{e.g.}, template, DLSCA) to characterize the leakage model related to the logits' manipulation.
\end{itemize}

These steps require a strong assumption that we will detail in \hyperlink{subsection.4.1}{Section 4.1}. Once this step is performed, our global attack methodology can be illustrated as in \autoref{fig:attack_phase}.
It follows the same general steps as a classical ZOO or SPSA attack by querying the network with an input and transforming progressively the image until an adversarial example is created as described in \hyperlink{subsection.2.4}{Section 2.4}. 
This is an iterative process with one iteration being composed of the following steps:
\begin{itemize}
\item (1) One data is sent $n$ times to the embedded neural network for prediction;
\item (2) Meanwhile, the attacker captures the $n$ corresponding side-channel traces containing information related to the manipulated logits;
\item (3) The attacker performs the logit extraction attack using a sub-optimal solution to \hyperlink{proposition.1}{Proposition 1}, \textit{e.g.}, templates or DNN models (see \hyperlink{subsection.2.5}{Section 2.5)}, and the newly captured attack traces. Then, the extracted logits are provided to the adversarial example methods;
\item (4) Using a gradient free method (\textit{e.g.}, ZOO, SPSA), the attacker estimates the perturbation to add to the inputs in order to create a successful adversarial example that will fool the embedded neural network.
\end{itemize}

Since a single iteration is usually not sufficient to construct a successful adversarial example, steps (1) to (4) are repeated until the perturbation causes a misclassification from the network (\textit{i.e.}, step (5) in \autoref{fig:attack_phase}) or until the generation framework stops.
Therefore, if we denote $m_{\text{adv}}$ as the maximum number of iterations to create an adversarial example and $n$ the maximum number of traces necessary to successfully extract the logits, the global complexity $C$ of the attack can be written as
\begin{equation} \label{eq:complexity}
\begin{split}
C &= \mathcal{O}(n \cdot m_{\text{adv}}).
\end{split}
\end{equation}

In comparison with gradient free optimization methods, the overall attack complexity is multiplied by the number of attack traces that are needed to retrieve the logits. 
But it is an overall improvement, as it allows attackers to perform an end-to-end black box attack.
However, as shown in \autoref{eq:complexity}, the practicability of the attack scenario highly depends on the ability of the side-channel attack (\textit{i.e.}, template attacks, DLSCA) to extract the logits.

\subsection{General Observation}\label{logit_extract_subsec}

\paragraph{Logits extraction.}
As mentioned in the previous section, the main idea in our attack consists in retrieving the logits manipulated by the embedded neural network through the use of side-channel attacks.
In particular, the targeted operations induced in a layer, or in an activation function, (as illustrated in \hyperlink{subsection.2.2}{Section 2.2}) should probably leak information on the logits.
Since the softmax function is computed to approximate the probability distribution $\texttt{Pr}[Y|X=x]$ from the logits, it is a natural target from a side-channel perspective to extract information related to these sensitive variables.
In addition, as side-channel attacks are still new to DL framework, no countermeasures are implemented to protect the device against these threats. 
As our attack is independent from the architecture, since we target a natural and frequently used function, namely softmax, the proposed methodology can be applied to all neural network architectures which solve a classification task. 
Indeed, extracting the logits from the softmax is beneficial to perform gradient-free generation frameworks proposed in the state of the art.

\begin{figure}[b!] 
\begin{subfigure}{0.23\textwidth}
\includegraphics[width=\linewidth]{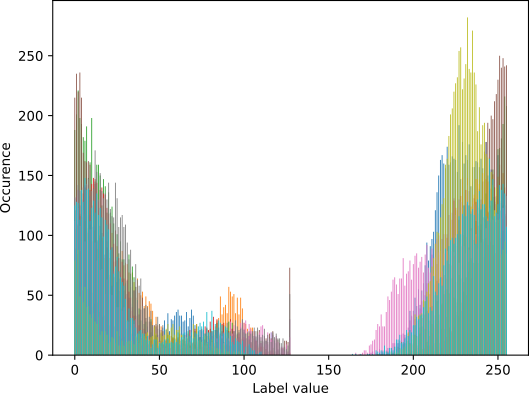}
\caption{Non-uniform distribution} \label{fig:non_uniform}
\label{fig:logits_distrib_non_uniform}
\end{subfigure}\hspace*{\fill}
\begin{subfigure}{0.23\textwidth}
\includegraphics[width=\linewidth]{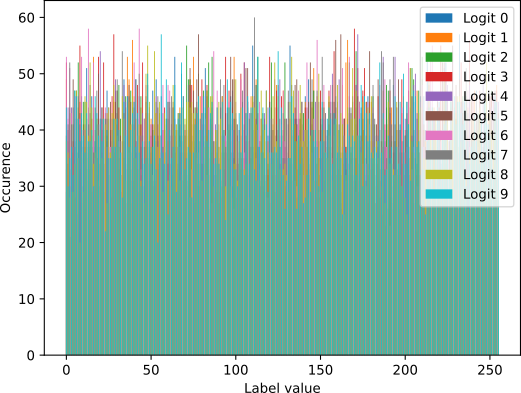}
\caption{Uniform distribution} \label{fig:uniform}
\label{fig:logits_distrib_uniform}
\end{subfigure}
\caption{Example of the distribution of logits on a real use-case over a set of $10000$ samples.}
\label{fig:logits_distrib}
\end{figure}

\paragraph{A natural distribution difficulty.}
As stated in \hyperlink{subsection.2.5}{Section 2.5}, template attack and DLSCA can be considered to extract the corresponding logits in a profiled scenario.
However, for an unbiased SCA, the profiling phase must be conducted with the logits uniformly distributed over $\mathcal{Y}$. 
The under-representation of certain logits will lead to poor estimation of the distribution parameters (\textit{i.e.}, the pair $(\boldsymbol{\mu}_{y}, \Sigma_{y})$ for template attack, and the weights for DLSCA) leading to a deterioration of the attack performance.
This issue is especially true when targeting the logits of a DNN.
Indeed, the purpose of the embedded neural network is to select an output among the possible classes in $\mathcal{Y}$, as such the DNN is trained to maximize one value in the logits vector and keep the other ones as low as possible.
It is then likely that, for a given model, the distribution of the logits will be over-represented for the lower values of the output range and under-represented in the higher part.
An example of class distributions is provided in \autoref{fig:logits_distrib_non_uniform}. The logits, in this example, are represented using 8-bit signed integers, as such the values over 127 correspond to negative logits. We can then observe that the distribution follows the pattern of over and under-representation of classes that we previously described. 
If the adversary targets an open-source DL framework (\textit{e.g.}, PyTorch \cite{pytorch_2019}, Tensorflow \cite{Tensorflow_2016}, FINN \cite{FINN_2017},  NNOM \cite{jianjia_ma_2020_4158710}), the source code can be modified in order to force the uniform distribution of the logits during the profiling phase.
Because most of the embedded neural networks are generated from open-source DL frameworks, the attacker can naturally have an access to the source code.
Then, the attacker can modify this code on the open device $\mathcal{D}$ to mitigate this issue, as illustrated in \autoref{fig:logits_distrib_uniform}. Added to the fact that our attack is architecture-agnostic, this allows the malicious user to build an unbiased SCA which can be applied on all the targeted embedded devices $\mathcal{D}^*$ using the same open source DL framework as in $\mathcal{D}$. 

This strategy has been considered in this paper and the following section proposes to experimentally validate this new attack strategy.



%% file: experiment.tex
\section{Experimental results}
In this section, we apply our methodology on an neural network implemented in a microcontroller.
First, we define the threat model that we consider. Then, \hyperlink{subsection.4.2}{Section 4.2} describes the experimental setup, while \hyperlink{subsection.4.3}{Section 4.3} details the security flaw exploited in this paper. Finally, \hyperlink{subsection.4.4}{Section 4.4} (resp. \hyperlink{subsection.4.5}{Section 4.5}) evaluates the practicability of the logits extraction (resp. the end-to-end attack scenario).

\subsection{Threat Model}
\paragraph{Scenario.}
In this study we contextualize the threat model we consider in this paper.
Following \hyperlink{section.3}{Section 3}, the primary goal of the attackers is to extract the logits from the embedded neural network by performing side-channel attacks.
In this paper, our strategy is based on profiling attack scenario, namely template attacks and DLSCA.
As mentioned in \hyperlink{subsection.2.5}{Section 2.5}, such scenario suggests that the attacker has a physical access to a copy of the targeted device.
This threat considers attackers who obtain the same hardware reference $\mathcal{D}$ as the targeted device $\mathcal{D}^{*}$ on the market.
As it is assumed that the DL framework used to embed the DNN model on the targeted device is known, the attacker can build its own model on $\mathcal{D}$ in order to extract the logits from $\mathcal{D}^{*}$.
Indeed, as the attack scenario introduced in \hyperlink{section.3}{Section 3} is architecture-independent (\textit{i.e.}, the softmax function is targeted), the knowledge of the DNN structure $\mathcal{D}^{*}$ is not required.
Consequently, the attacker can easily construct its own copy on $\mathcal{D}$ of the targeted function embedded in $\mathcal{D}^{*}$ to satisfy the requirements of the profiling attack scenario.
Once this profiling phase has been performed on $\mathcal{D}$, the attacker can extract the logits from the targeted device $\mathcal{D}^{*}$.
Therefore, this threat model is considered at inference stage.
In other words, the targeted embedded neural network has already been trained and deployed.
For example, this threat considers attackers who come from the DNN accelerator design team, or as insiders in the companies hosting DNN infrastructure so that they are capable to get a physical access to the device. 

\paragraph{Capability.}
To conduct such scenario, it is assumed that the attacker has knowledge on the DL framework embedded into the targeted system.
Knowing the reference of the targeted hardware device, the adversary can modify the source code of the DL framework to mitigate the issues introduced in \hyperlink{subsection.3.2}{Section 3.2}.
This assumption is in accordance with the scenario previously introduced.
Secondly, attackers can acquire power consumption or electromagnetic emanation traces of the targeted embedded neural network in high resolution through the use of an oscilloscope.
This assumption is in accordance with the scenario previously introduced.
As the attack strategy targets a function of the DL framework that is independent of the underlying neural network architecture (\textit{e.g.}, softmax function), no prerequisite on the DNN structure is needed.

\subsection{Experimental Setup}\label{subsec:experimental_setup}
\paragraph{Hardware implementation.}
We validate our contribution by targeting an ARM Cortex-M7 microcontroler unit (MCU) on an STM32-F767 board running at a frequency of 216 MHz. This board incorporates 2 Mbytes of flash memory and 512 Kbytes of SRAM, which is enough to mount quantized models on the board. We choose to target a classifier based on the denseNet \cite{Huang2016DenselyCC} architecture, including a softmax as the last layer. The model was trained on the MNIST dataset with the Adam optimizer, and using Tensorflow \cite{Abadi2016TensorFlowAS} for development. MNIST is a dataset for image recognition on handwritten digits composed of $60,000$ training data and $10,000$ test images~\cite{726791}. This dataset considers $10$ classes. 
As a consequence, $10$ logits must be extracted to validate the methodology introduced in \hyperlink{section.3}{Section 3}.
In the context of embedded system, we choose to use quantization to reduce the impact on the resources. The inputs, weights, and activation values of the model are quantized to $8$ bits, using deterministic PTQ. NNOM, an open-source framework, was used to facilitate the incorporation on the MCU~\cite{jianjia_ma_2020_4158710}. NNOM translates the model from a Tensorflow object to a C project, which can then be directly build and run on the board. The generated project was compiled without optimisation from the compiler. We decided to only test on the NNOM framework but we expect our attack methodology to translate to other frameworks, as any unprotected manipulation of the logits in the assembly code will induce side channel leaks. We leave evaluation of other frameworks and hardware implementations for future work.   

\begin{figure}
    \centering
    \includegraphics[width=0.4\textwidth]{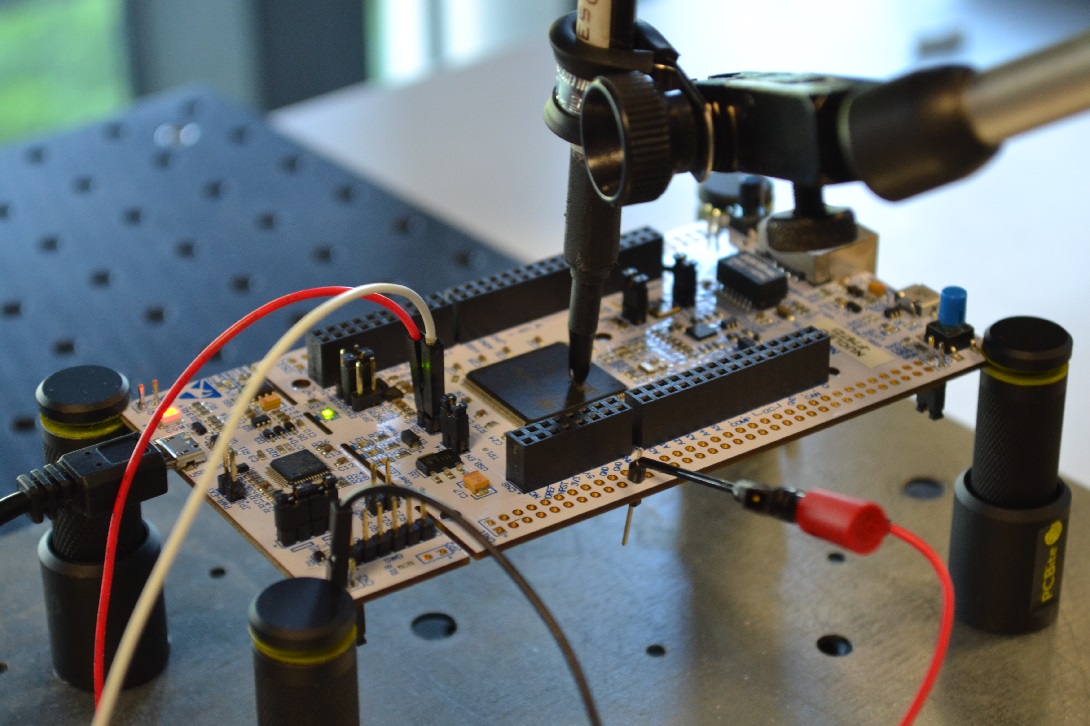}
    \caption{Experimental setup.}
    \label{fig:experimental_setup}
\end{figure}

\paragraph{Practical setup.}
A probe from Langer (EMV-Technik RF-U 2,5 with a frequency range going from 30MHz up to 3GHz) is connected to an amplifier (ZLF-2000G+) to capture the electromagnetic (EM) side-channel leakages. 
To acquire the EM signal, a Lecroy oscilloscope (2.5 GHz WaveRunner 625Zi) is used in our setup. 
To ease the leakage exploitation, a trigger surrounding the targeted function (\textit{i.e.}, \texttt{local\_softmax\_q7()} from NNOM) has been added.
As the softmax computation is distinguishable from the other patterns, the configuration of the trigger is not considered difficult.
A quick overview of our experimental setup is illustrated in \autoref{fig:experimental_setup}.

\begin{table*}[t]
        \resizebox{\textwidth}{!}{%
        \begin{tabular}{|c|c|c|c|c|c|c|c|c|c|c|}
            \hline
            & Logit 0 & Logit 1 & Logit 2 & Logit 3 & Logit 4 & Logit 5 & Logit 6 & Logit 7 & Logit 8 & Logit 9\\
           \hline
           Maximum SNR value & $2.3 \cdot 10^{-1}$ & $2.2 \cdot 10^{-1}$ & $5.0 \cdot 10^{-2}$ & $3.5 \cdot 10^{-2}$ & $2.0 \cdot 10^{-2}$ & $2.0 \cdot 10^{-2}$ & $1.5 \cdot 10^{-2}$ & $1.5 \cdot 10^{-2}$ & $1.6 \cdot 10^{-2}$ & $1.0 \cdot 10^{-1}$ \\
           \hline 
           SNR threshold & $5.0 \cdot 10^{-3}$ & $5.0 \cdot 10^{-3}$ & $8.0 \cdot 10^{-3}$ & $8.0 \cdot 10^{-3}$ & $8.0 \cdot 10^{-3}$ & $5.0 \cdot 10^{-3}$ & $5.0 \cdot 10^{-3}$ & $3.5 \cdot 10^{-3}$ & $3.5 \cdot 10^{-3}$ & $3.5 \cdot 10^{-3}$ \\
           \hline 
           Dimension of reduced traces & $6,514$ & $8,840$ & $1,654$ & $1,302$ & $774$ & $2,866$ & $2,223$ & $5,813$ & $6,186$ & $6,299$ \\
           \hline 
        \end{tabular}
        }
    \caption{Preprocessing on the EM traces depending on the obtained SNR value.}
    \label{tab:comparaison}
\end{table*}
To get a better insight on our contribution, $10$ independent attacks are performed and the success rate metric is provided to identify the number of attack traces that are required to successfully retrieve each logit.
As a reminder, the success rate metric defines the probability that an attack succeeds in recovering the true logit amongst all hypotheses in $\mathcal{Y}$.

\paragraph{Logits distribution.}
Following the recommendations provided in \hyperlink{subsection.3.2}{Section 3.2}, we evaluated the distribution of the logits on the trained embedded neural network. As expected, a non-uniform distribution is observed (see \autoref{fig:logits_distrib_non_uniform}). 
To solve this issue, we modified the source code of the NNOM softmax function to force its input (\textit{i.e.}, the logits) to follow a uniform distribution.
The results we obtained are highlighted in \autoref{fig:logits_distrib_uniform}.
This requirement is highly recommended to mitigate the bias in the suboptimal solution of \hyperlink{proposition.1}{Proposition 1} (\textit{i.e.}, template attacks, DLSCA), but not mandatory as stated in \hyperlink{subsection.3.2}{Section 3.2}. 

\subsection{Security Flaw}\label{subsec:security_flaw}
To validate the methodology introduced in \hyperlink{subsection.3.1}{Section 3.1}, the NNOM softmax function is targeted.
We choose this implementation as it is a common solution in the literature \cite{zinng_nnom, Rusci_2023, Joud_CARDIS_2023, dumont2023evaluation}.
Whereas traditional softmax function is defined by a normalization which maps the logits (or the confidence scores) to a probability distribution following $s(\textbf{Z}[i]) = \frac{e^{\textbf{Z}[i]}}{\sum_{j=1}^{|\mathcal{Y}|}e^{\textbf{Z}[j]}}$ where $\textbf{Z}[i]$ denotes the logits related to the $i^{\text{th}}$ class, the NNOM framework proposes an alternative to this natural approach.
In particular, instead of computing the exponentiation of each logit as suggested in the classical softmax function, NNOM captures the maximum value $\texttt{max}_{z}$ from the vector of logits $\textbf{Z}$.
Then, all logits lower than $(\texttt{max}_{z} - 8)$ are ignored, while others are normalized following the function $\texttt{local\_softmax\_q7}()$.
In particular, \autoref{lst:softmax_c} is an extract of the C code source of the softmax computation from NNOM.
This portion searches the maximum value from the $\texttt{vec\_in}$ vector which characterizes the vector of logits $\textbf{Z}$.

\begin{figure}
    \centering
    \begin{subfigure}{0.15\textwidth}
\begin{lstlisting}[caption=Part of the \texttt{local\_softmax\_q7} implementation in NNOM., 
            label={lst:softmax_c}, 
            language=C, 
            backgroundcolor=\color{backcolour},   
            commentstyle=\color{codegreen},
            keywordstyle=\color{magenta},
            numberstyle=\tiny\color{codegray},
            stringstyle=\color{codepurple},
            basicstyle=\ttfamily\footnotesize,
            breakatwhitespace=false,         
            breaklines=true,                 
            keepspaces=true,                 
            numbers=left,       
            numbersep=5pt,                  
            showspaces=false,                
            showstringspaces=false,
            showtabs=false,                  
            tabsize=2,
            frame=single, 
            firstline=9, 
            lastline=17]
	/* Base is initialized to (int16_t)-257 */
    /* We first search for the maximum */
    for (i = 0; i < dim_vec; i++)
    {
        if (vec_in[i] > base)
        {
            base = vec_in[i];
        }
    }
\end{lstlisting}
    \end{subfigure}
    \hfill
    \begin{subfigure}{0.25\textwidth}
\begin{lstlisting}[caption=Assembly code of line 5 from Listing (1)., 
            label={lst:assembly_code}, 
            language={[x86masm]Assembler}, 
            morekeywords={b.n, ldr, ldrsb.w, sxth, ldrsh.w, bge.n, strh, adds, bhi.n},
            backgroundcolor=\color{backcolour},   
            commentstyle=\color{codegreen},
            keywordstyle=\color{magenta},
            numberstyle=\tiny\color{codegray},
            stringstyle=\color{codepurple},
            basicstyle=\ttfamily\footnotesize,
            breakatwhitespace=false,         
            breaklines=true,                 
            keepspaces=true,                 
            numbers=left,       
            numbersep=5pt,                  
            showspaces=false,                
            showstringspaces=false,
            showtabs=false,                  
            tabsize=2,
            frame=single]
; if (vec_in[i] > base)
ldr	r3, [r7, #32]	; Loading the address of the index "i" 
ldr	r2, [r7, #12]	; Loading the address of z 
add	r3, r2 			; Set the pointer to the i-th element of confidence score z
ldrsb.w	r3, [r3]	; Loading of the i-th element of z 
sxth	r3, r3		; Extension the i-th element of z to a 32-bit 
ldrsh.w	r2, [r7, #30] ; Base value loading
cmp	r2, r3			; Comparison between base and the i-th element of z
\end{lstlisting}
    \end{subfigure}
    \caption{C and Assembly codes from the NNOM softmax function.}
    \label{fig:softmax_source_code}
\end{figure}

For each element included in $\texttt{vec\_in}$, the base value is updated if its value is lower than the $i^{\text{th}}$ element of $\texttt{vec\_in}$ (\textit{i.e.}, $\textbf{Z}[i]$).
\autoref{lst:assembly_code} defines the assembly code which characterizes the ``if''-condition involved in line 5 of \autoref{lst:softmax_c}. 
The logit $\textbf{Z}[i]$ stored into the Flash memory is loaded into register $\texttt{r3}$ (see line 5 in \autoref{lst:assembly_code}).
During the load instruction, a side-channel attack can be performed to extract information related to the targeted logit.
In addition, the $\texttt{base}$ value is updated if the ``if''-condition is respected. 
Therefore, when the $\texttt{base}$ value is loaded into register $\texttt{r2}$ (line 7 in \autoref{lst:assembly_code}), leakages could be observed on the current maximum value in $\texttt{vec\_in}$.
This load instruction suggests the manipulation of some $z_{i}$ values multiple times which can make the leakage exploitation more practicable.
An noteworthy remark is that the side-channel attack considered in this paper can be easily transferred from a DNN architecture to another as long as \texttt{local\_softmax\_q7} of NNOM is used as well as the Nucleo-144.

\begin{figure}[b!] 
\centering
    \begin{subfigure}{.23\textwidth}
        \includegraphics[width=\linewidth]{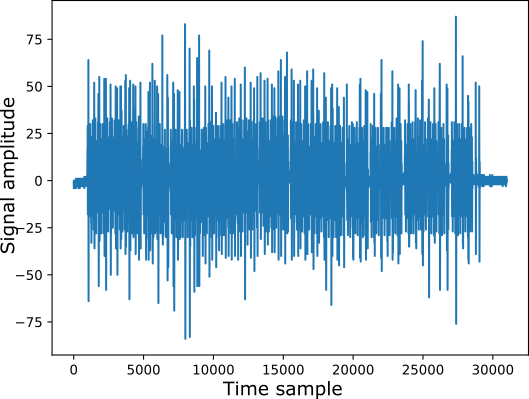}
        \caption{Electromagnetic trace of the softmax execution.}
        \label{fig:EM_softmax}
    \end{subfigure}%
    \hfill
    \begin{subfigure}{.23\textwidth}
        \includegraphics[width=\linewidth]{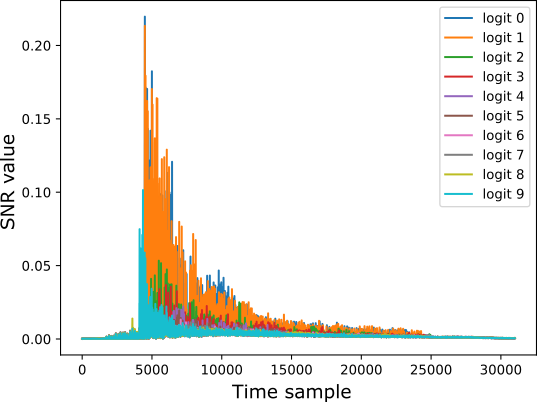}
        \caption{SNR associated with each logit.}
        \label{fig:SNR_logit0}
    \end{subfigure}
    \caption{Preprocessing of an EM trace.}
    \label{fig:example_preprocessing_logit0}
\end{figure}

We wanted to confirm our analysis of this security flaw by targeting directly for a signal associated with the value of logits. \autoref{fig:EM_softmax} represents the  electromagnetic trace associated to the processing of the softmax function for different logits. \autoref{fig:SNR_logit0} displays the SNR for the first logit computed. As expected from our analysis of \autoref{lst:softmax_c}, the ratio between the signal associated to this variable and the noise change during the execution time. This displays the manipulation of the targeted variable in the assembly code. Remark: readers have to notify that the scenario proposed in this paper can be conducted on any type of DL implementation since the manipulation of the logits occurs in the assembly code.  

To validate our attack methodology, for each logit, a set of $1,900,000$ EM traces of $31,000$ time samples is captured for the profiling set, while $1,000$ traces are acquired for the attack phase. 
To reduce the needs of computation, the EM traces are reduced following the SNR peaks we obtained on each logit.
In that purpose, a threshold is configured for each logit such that all the time samples associated with a higher SNR peak are kept.
The threshold is defined in order to find the best trade-off between computational time and leakage information.
The results we obtained are defined in \autoref{tab:comparaison}.
As mentioned in \hyperlink{subsection.2.5}{Section 2.5}, other dimensionality reduction techniques such as PCA \cite{Archambeau_CHES_2006}, LDA \cite{Choudary_CARDIS_2015} or KDA \cite{Cagli_CARDIS_2017} could be used as suitable alternatives.

\subsection{Evaluation of the Logit Extraction}\label{subsec:attack_logits}
To evaluate the logit extraction process, three attacks have been conducted. All attacks have the same number of profiling traces on the open sample $\mathcal{D}$ to build their sub-optimal solution to \hyperlink{proposition.1}{Proposition 1}. On the contrary, the number of attack traces used on the targeted sample $\mathcal{D}^*$ to achieve a similar average success rate, is different from one attack to another. This metric will be used to compare the efficiency of the three methods.
First, template attacks \cite{ches-2002-635} are performed in order to assess the ability of Gaussian distribution to capture the dependencies between the targeted logits and the EM traces.
As such assumption can not hold in practice, we evaluated a more generic solution which considers a broad family of models: \textit{multinomial logistic regression}.
As suggested by Masure \textit{et al.} \cite{Masure_ePrint_2023}, this class of models defines the set of all polynomial transformations such that a clear connection can be proposed with template attacks.
Finally, to get a full overview of the methodology, MLPs are constructed in order to perform a deep learning-based side-channel attack and recover the logits of the embedded neural network.
While the latter solution is beneficial from a performance perspective, the lack of interpretability makes its configuration a difficult task.

\begin{figure*}[t!] 
\begin{subfigure}{0.32\textwidth}
\includegraphics[width=\linewidth]{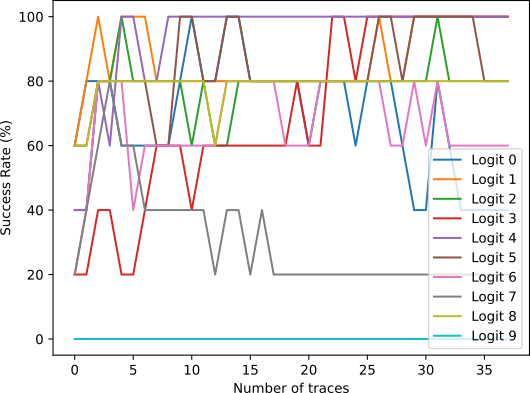}
\caption{Template attacks} \label{fig:result_TA}
\end{subfigure}\hspace*{\fill}
\begin{subfigure}{0.32\textwidth}
\includegraphics[width=\linewidth]{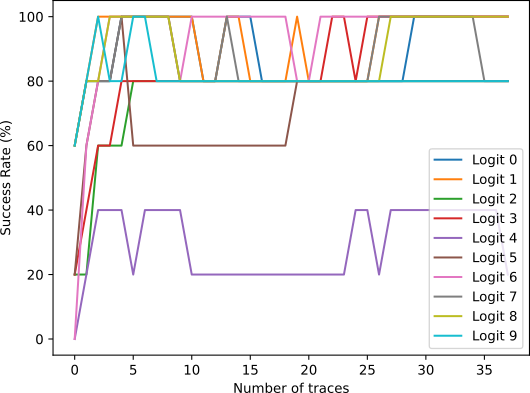}
\caption{Multinomial logistic regression} \label{fig:result_LR}
\end{subfigure}\hspace*{\fill}
\begin{subfigure}{0.32\textwidth}
\includegraphics[width=\linewidth]{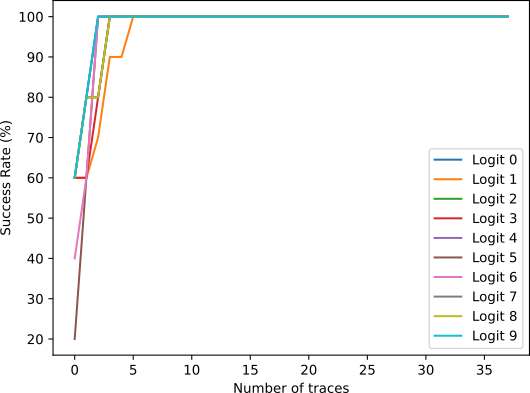}
\caption{Deep-learning based side-channel attacks} \label{fig:result_DLSCA}
\end{subfigure}
\caption{Success rate results on targeting NNOM softmax function. Each colour of a curve corresponds to the success rate of a given logit (out of $10$) as a function of the number of traces used for the attack.}
\label{fig:result_attack}
\end{figure*}

\paragraph{Template attack results.}
The template attacks are the naive solution when profiled attacks are conducted.
The obtained results are displayed in \autoref{fig:result_TA} where each colour of a curve corresponds to the success rate for a given logit (out of $10$) as a function of the number of traces used for the attack.
Accordingly, the purple curve corresponds to the success rate for $5^{\text{th}}$ logit out of the $10$ possible classes averaged on $10$ independent experiments.
For this specific logit, $8$ attack traces are required to successfully extract (\textit{i.e.}, a success rate of $100\%$) the targeted logit value. 
From \autoref{fig:result_TA}, it can be also observed that after only $11$ attack traces, the average attack success rate is around $66\%$.
In addition, while some logits are recovered with a very small amount of traces (\textit{e.g.}, $8$ traces for the $5^{\text{th}}$ logit), others are not successfully retrieved (\textit{e.g.}, no successful attack is obtained when the $10^{\text{th}}$ logit is targeted).
This observation suggests that additional investigations can be conducted to improve even more the performance of template attacks.
An attacker could increase the number of collected traces on the open sample $\mathcal{D}$ in order to improve the profiling phase and build a better solution to \hyperlink{proposition.1}{Proposition 1}. Other setup configurations (\textit{e.g.}, dimensionality reduction technique, SNR threshold) could significantly improve the performance gain.
The results we obtained illustrate the practical limitation of the template attacks.
To mitigate this issue, an attacker can use another technique based on multinomial logistic regression in order to consider a less restrictive class of model.

\paragraph{Multinomial logistic regression results.}
To investigate the benefits of this class of models, a simple MLP without hidden layers, nor activation functions, is configured.
Only an output softmax function is used to solve the multiclass-classification problem.
During the profiling phase we used the traces collected on $\mathcal{D}$ and the associated labels (\textit{e.g.} the logits) to train the MLP via supervised learning using the Adam optimizer \cite{Kingma_2014}.
This profiling process is conducted over $20$ epochs such that the batch size (resp. the learning rate) is set to $512$ (resp. $10^{-5}$). 

In comparison with the template attack result, \autoref{fig:result_LR} shows a performance improvement when the multinomial logistic regression is considered.
In particular, after $10$ attack traces, the average success rate is about $80\%$, and with this method all the logits are extracted at least once, which was not the case with the template attack.
However, the attack on the $5^{\text{th}}$ logit performs poorly in comparison with other logits.
Consequently, if an attacker wants to follow the methodology introduced in \hyperlink{section.3}{Section 3}, the wrong label estimation will impact the generation of adversarial examples.
Even if some state-of-the-art results suggest that the knowledge of the top-$k$ logits is sufficient to generate adversarial examples \cite{pmlr-v80-ilyas18a,ilyas2018prior}, this scenario is not considered in the scope of this paper.

\paragraph{DLSCA results.}
Finally, the most generic solution consists in designing a DNN model to approximate the best possible solution to \hyperlink{proposition.1}{Proposition 1}. We target the embedded system developed to solve the classification problem on the MNIST dataset running on the targeted device $\mathcal{D}^*$, with another network trained on the EM traces of the logit's processing on the open device $\mathcal{D}$.
The MLP trained on the traces of $\mathcal{D}^*$ is composed of $3$ hidden layers of $1,000$, $1,000$ and $100$ neurons respectively and, in order to approximate non-linear functions, each neuron uses the ReLU activation function.
The same configuration as for the multinomial logistic regression is used (\textit{i.e.}, $20$ epochs and a batch size of $512$), whereas the learning rate is adapted to $10^{-4}$.
The results obtained with this setting are observed in \autoref{fig:result_DLSCA}.
Interestingly, when the DL technique is used to extract the logits from the embedded neural network, all the sensitive variables are retrieved within $5$ EM traces. As expected, this method offers the best performance compared to the others.  \\

Through this process, it can be confirmed that an attacker can perform side-channel attacks using a sub-optimal solution to \hyperlink{proposition.1}{Proposition 1} to extract the logits from the embedded neural network.
Regarding the global complexity of the attack ($C$ in \autoref{eq:complexity}), the attacker needs $5 \times m_{\text{adv}}$ attack traces to create a given adversarial example. 
In our experiment, we set $m_{\text{adv}}$, the maximum number of iterations to create an adversarial example, to $10,000$.
Consequently, if the generation of adversarial examples is possible, the total number of attack traces required to conduct an end-to-end attack is $50,000$ in the worst case\footnote{In practice, to defeat a protected cryptographic system, a side channel attack may require millions of traces.}.
However, it should be mentioned that, sometimes, $10,000$ iterations are not sufficient to generate such adversarial example using the ZOO framework. 
Consequently, the total number of attack traces highly depends on the performance of the adversarial example tool.
The following section investigates an end-to-end scenario of our proposition through the use of the ZOO framework.

\subsection{Evaluation of the Complete Attack}\label{subsec:end2end_attack}

Once the extraction is performed using one of these methods, an attacker can perform the generation of adversarial examples in a black box setting using gradient-free frameworks.
To validate this statement, we use the open-source implementation of the ZOO attack\footnote{\url{https://github.com/IBM/ZOO-Attack}} provided by \cite{Chen2017ZOOZO} to attack our embedded model. Since this framework uses the probability vector in the objective function, as described in \hyperlink{subsection.2.4}{Section 2.4}, we compute it with the extracted logits and solve the optimization problem accordingly. We confront the results of our attack with those obtained after white-box generation using the Basic Iterative Method \cite{Kurakin2016AdversarialEI} from the Cleverhans framework \cite{papernot2018cleverhans} on a substitute network. The substitute network is based on the same architecture as the targeted network and we use the same methodology as \cite{Rakin2021DeepStealAM} and trained it using 10\% of the training data labeled by the embedded model.
We evaluate the transfer rate, the average distortion, and the average number of request, to generate an adversarial example. Our results are included in \autoref{tab:Result_full}. 

The number of request necessary to produce an adversarial example is much higher when using our framework. This is caused by the profiling attack which needs requests to the system to acquire the physical traces and build the model. The number of traces for this phase is highly dependent on the implementation and the surrounding noise. Since we are using optimization method in our profiling phase, this number is highly experimental and could be much lower for another hardware. However, the main advantage of our framework is its genericity. Once the profiling phase is finished, an attacker will not have to fully retrain it while targeting any models implemented using NNOM. On the other hand, to train the substitute network, the malicious user will need to have access to training data specific for each task and will potentially have to retrain a substitute for each new targeted DNN.  As expected, even in an optimal case with a substitute based on the same architecture, not all the white-box-generated adversarial examples are transferable with an average rate of 56\%. Since ZOO directly uses the targeted network, the generated examples are by definition directly adversarial for this network. The high value in the distortion for both frameworks seems to be due to the 8-bit quantization of the input, which forces the perturbation to be larger to cause an impact. \\


\begin{table}[t!]
        \resizebox{\columnwidth}{!}{%
        \begin{tabular}{|c|c|c|c|}
            \hline
            & Transfer rate (\%) & Average Distortion (L2) & Number of request\\
           \hline
           Logits extraction and ZOO & $ 100 $ & $7.54$ & $ ~1.9 \cdot 10^{6}$\\
           \hline 
           Substitute network & $56.61$ & $5.80$ & $6000$ \\
           \hline 
        \end{tabular}
        }
    \caption{Comparison between our framework and generation on a substitute network.}
    \label{tab:Result_full}
\end{table}


With this design, we demonstrated that an attacker can extend existing state-of-the-art frameworks to generate powerful adversarial examples in a black box scenario using our end-to-end methodology. 
We only evaluated our framework with the ZOO attack and against only a substitute network, as they are no other method, to the best of our knowledge, that proposes a generation process that does not require access to any internal or output values (logits, confidence score). In that context, the purpose of this paper is to introduce a method to reduce the necessary assumptions. The case of boundary attacks \cite{Brendel2017DecisionBasedAA}, which only use the output labels to generate adversarial example, can even be considered for situations where these labels are not directly known by the attacker. We leave the comparison with boundary attacks, when access to the output labels is possible, for future work, as well as evaluations on more complex datasets.

%% file: conclusion.tex
\section{Conclusion}

This study has for purpose to prove that it is worth considering side-channel and, more generally, physical attacks against DL implementation, in addition to software attacks. To do so, we demonstrate the ability of an attacker to generate powerful adversarial examples by breaking the black box assumption through the exploitation of side-channel attacks against an embedded DNN. 
In particular, we contribute to the state of the art by adding a side-channel extraction of the logits to gradient-free adversarial example frameworks using these values.

To validate our methodology, a practical scenario has been conducted on a microcontroller implementing a denseNet with the NNOM framework.
Based on a code analysis, security flaws have been identified and exploited in the NNOM softmax function and the logit vector has been extracted, in the best case, within 5 attack traces. We finalize our attack by combining our physical attack with the ZOO framework for adversarial example generation, confirming the complementarity between software and hardware attacks. 

To further improve our contribution, some future works can be suggested. Vulnerability in the softmax function can be studied in some widely used open-source DL frameworks (\textit{e.g.}, PyTorch, Tensorflow, FINN) or on more optimized hardware implementations (\textit{e.g.}, on FPGA). Further investigation on the attack can be focused on the impact of a wrong logit estimation on the final attack and on adapting it to perform non-profiled side-channel attacks, which reduce the attacker's capability. Finally, while the security of embedded AI system is still a new research direction, finding countermeasures against our contribution is crucial to make the AI solutions more secure.